\documentclass{sig-alternate-05-2015}
\usepackage{hyperref}
\usepackage{psfrag}
\usepackage{epstopdf}
\usepackage{graphicx}
\usepackage{subfig}
\usepackage{times}
\usepackage{amsmath} 
\usepackage{amssymb}  
\usepackage{mathabx}
\usepackage{mathptmx}
\usepackage{color}
\usepackage{enumitem}
\usepackage{multirow}

\setlist[itemize]{noitemsep, topsep=1pt}
\setlist[enumerate]{noitemsep, topsep=1pt}
\setlist[description]{noitemsep, topsep=1pt}

\begin{document}
\sloppy

\CopyrightYear{2016}

\setcopyright{acmcopyright}

\conferenceinfo{ICDSC '16,}{September 12-15, 2016, Paris, France}

\isbn{978-1-4503-4786-0/16/09}\acmPrice{\$15.00}

\doi{http://dx.doi.org/10.1145/2967413.2967435}





%

\title{3D Reconstruction with Low Resolution, Small Baseline and High Radial Distortion Stereo Images}
%
%
%
%
%

\numberofauthors{2} 
%
\author{
%
%
  \alignauthor
  Tiago Dias and Helder Araujo\titlenote{\small{T. Dias and H. Araujo were 
  partially funded by QREN Programme ''Mais Centro'', Grant SCT 2011 
  02 027 4824}}\\
  \affaddr{Institute of Systems and Robotics,}\\
  \affaddr{University of Coimbra}\\
  \affaddr{Pinhal de Marrocos -- Polo II,\\3030 Coimbra - Portugal}\\
  \email{\normalsize \tt \{tdias\&helder\}@isr.uc.pt}
  \alignauthor
  Pedro Miraldo\titlenote{\small{P. Miraldo was partially funded with grant {\tt SFRH/BPD/111495/2015}, from {\it Funda\c{c}\~{a}o para a Ci\^{e}ncia e a Tecnologia}}}\\
  \affaddr{Institute for Systems and Robotics,}\\
  \affaddr{Instituto Superior T\'{e}cnico,\\Universidade de Lisboa}\\
  \affaddr{Av. Rovisco Pais, 1, 1049-001 Lisboa, Portugal}\\
  \email{\normalsize \tt pmiraldo@isr.tecnico.ulisboa.pt}
}      

\maketitle
\begin{abstract}
  In this paper we analyze and compare approaches for 3D 
  reconstruction from low-resolution (250x250), high radial distortion 
  stereo images, which are acquired with small baseline 
  (approximately 1mm). These images are acquired with the system 
  NanEye Stereo manufactured by CMOSIS/AWAIBA. These stereo cameras 
  have also small apertures, which means that high levels of 
  illumination are required. The goal was to 
  develop an approach yielding accurate reconstructions, with
  a low computational cost, i.e., avoiding non-linear numerical 
  optimization algorithms. In particular we focused on the
  analysis and comparison of radial distortion models. To perform the 
  analysis and comparison, we defined a baseline method based on 
  available software and methods, such as the Bouguet 
  toolbox~\cite{bouguet} or 
  the Computer Vision Toolbox 
  from Matlab. The approaches tested were based on the 
  use of the polynomial model of radial distortion, and on the 
  application of the division model. The issue of the center of 
  distortion was also addressed within the framework of the application 
  of the division model. We concluded that the division model with a 
  single radial distortion parameter has limitations.
\end{abstract}

%
%

\ccsdesc[500]{Computing methodologies~Camera calibration}
\ccsdesc[500]{Computing methodologies~Epipolar geometry}
\ccsdesc[300]{Computing methodologies~3D imaging}

%

%
%

%
%
\printccsdesc


\keywords{3D reconstruction; Stereo; Radial distortion}

\section{Introduction}

\begin{figure}[t]
  \centering
  \includegraphics[height=1.5in, width=2in]{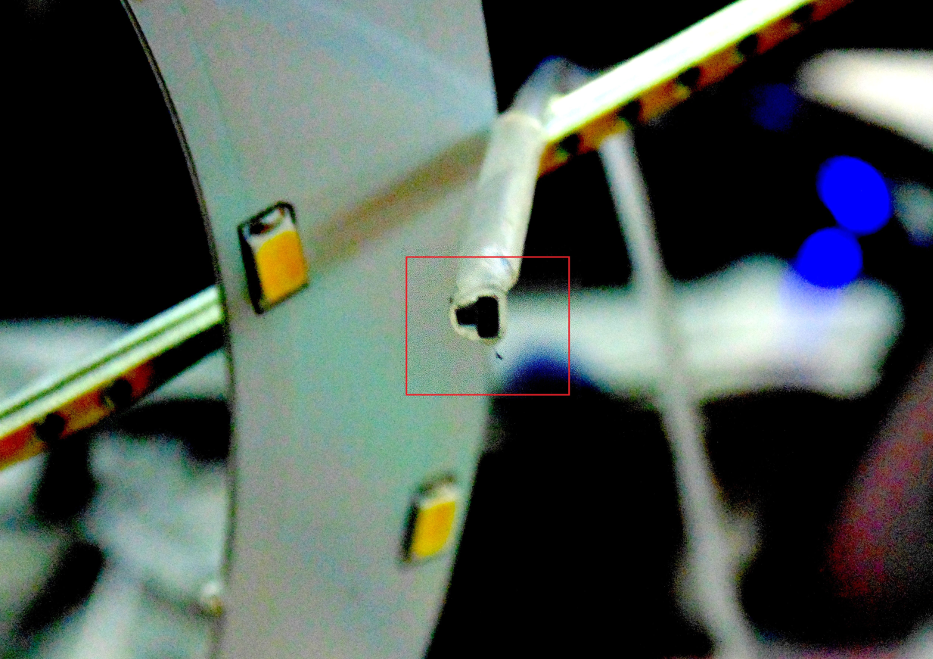}
  \caption{Image of the NanEye stereo pair.}
  \label{Fig:naneye}
\end{figure}

3D reconstruction has been subject of significant 
research~\cite{szeliski:2010}. 
Reconstruction from stereo pairs requires the calibration of both 
cameras, as well as of the system itself, namely of the relative pose 
between the cameras~\cite{hartley:2000}. The correspondences between 
the regions/features/pixels of left and right cameras allow the 
estimation of the 3D coordinates of the corresponding points. 
Algorithms for the 
establishment of correspondences have also been subject of intense 
research. The "Middlebury Computer Vision Pages" site contains 
several databases of stereo images for the evaluation of stereo 
algorithms~\cite{middlebury}, as well as results of the evaluations. 
In this paper we focus on modeling the NanEye stereo camera 
manufactured by 
CMOSIS/AWAIBA--see Fig.~\ref{Fig:naneye}. The main specifications of the 
system, 
as provided by the manufacturer, are shown in Tab.~\ref{Tab:1}
Our goal is the development of a model and approach so that 3D 
reconstructions with good accuracy can be performed despite the 
shortcomings of the system namely:
\begin{enumerate}
  \item Low resolution;
  \item High radial distortion; and
  \item Short baseline.
\end{enumerate}

\begin{figure}[t]
  \centering
  \includegraphics[height=2in, width=2in]{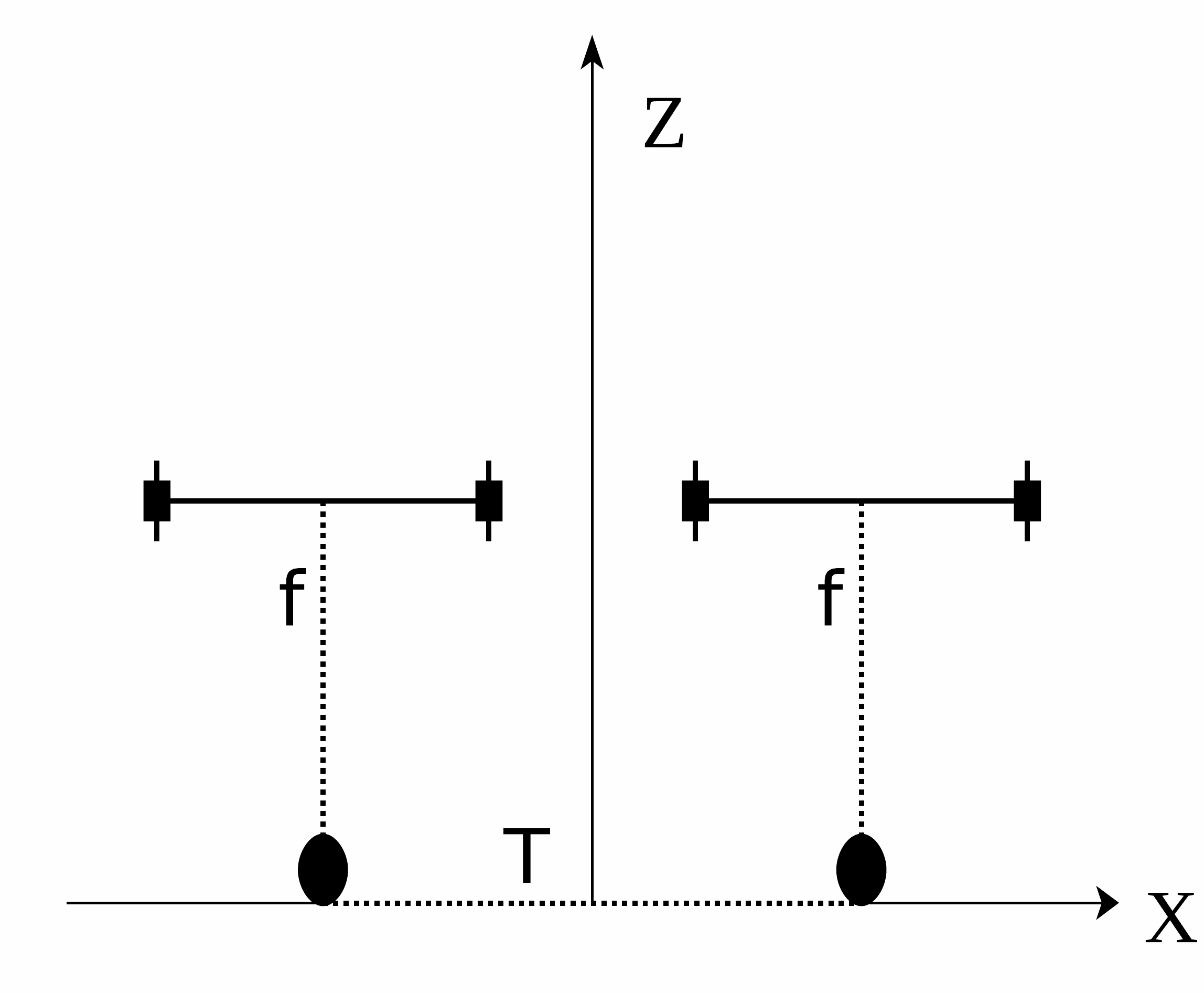}
  \caption{Representation of fronto-parallel stereo geometry.}
  \label{fig:fronto-parallel}
\end{figure}

Consider the geometry of a simple fronto-parallel stereo system 
(Fig.~\ref{fig:fronto-parallel}). In this configuration, $f$ is the 
focal length of both cameras, $T$ is the baseline length, $Z$ is the 
depth and $d$ is the image disparity. As it is well known and for 
this case disparity $d$ is given by:
\begin{equation}
  d=\frac{f T}{Z}
\end{equation}
and the uncertainty in depth $\delta Z$ is given by:
\begin{equation}
  \delta Z=\dfrac{Z^{2}}{f T} \delta d
\end{equation}
Therefore the short baseline length ($T$) implies that both small 
values and uncertainties in the disparity ($d$) will increase the 
uncertainties on the depth estimates.

We did not address the problem of estimating the stereo 
correspondences. This paper focuses on the issues of geometric 
modeling a stereo system with low resolution, high radial 
distortion, and short baseline, so that 3D reconstructions 
with good accuracy can be obtained.  

\begin{table}[t]
\centering
\small{
  \begin{center}
    \begin{tabular}{||l|l||}  \hline
      Resolution (pixels)	& $250\times 250$ \\ \hline
      Focal length ( $mm$ )	& $0.66$ \\ \hline
      F-number	& $2.7$ \\ \hline
      Pixel size ( $\mu m$ )	& $3\times 3$ \\ \hline
      Depth of focus ($mm$)	& $5.0-\varpropto$ \\ \hline
      Size ($mm$)	& $2.2 \times 1.0 \times 1.7$ \\ \hline	
    \end{tabular}
  \end{center}}
  \caption{Manufacturer's specifications of NanEye Stereo.}
  \label{Tab:1}
\end{table}

\section{Stereo System Geometry}
\begin{figure}[t]
  \centering
  \includegraphics[height=2in, width=3in]{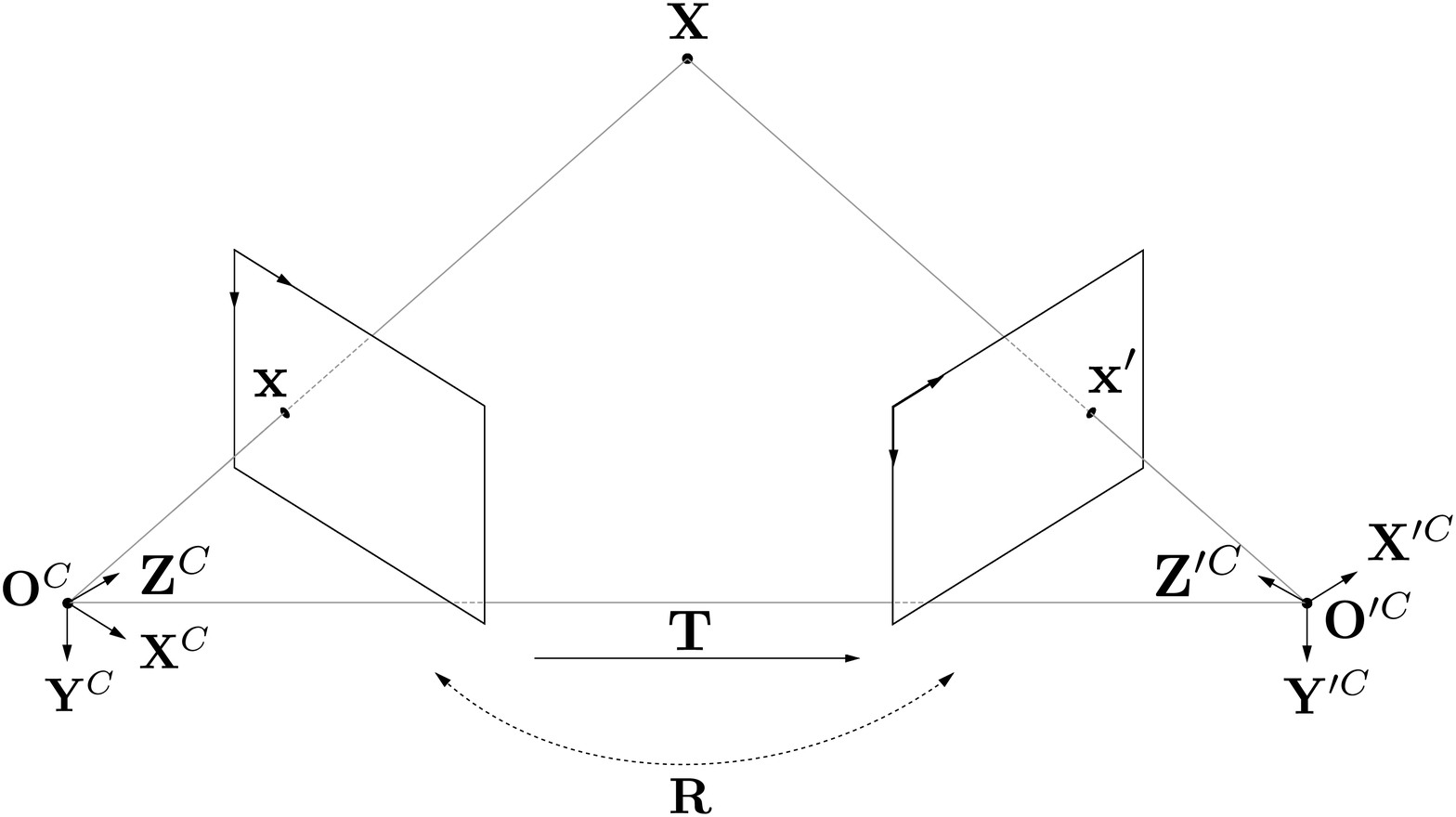}
  \caption{Generic representation of the stereo geometry.}
  \label{fig:stereo_system}
\end{figure}

The geometry of a generic stereo system can be represented as shown in 
Fig.~\ref{fig:stereo_system}.This figure is used 
for the purpose of definition and representation of the 
notations used in the paper, such that:
\begin{itemize}
  \item{$\mathbf{O}^C$ and $\mathbf{O}'^C$ represent the optical centers of the left and right cameras, respectively;}
  \item{$\mathbf{X}^C$, $\mathbf{Y}^C$, $\mathbf{Z}^C$ and $\mathbf{X}'^C$, $\mathbf{Y}'^C$, $\mathbf{Z}'^C$ represent the left and right camera coordinate systems, respectively; and}
  \item{$\mathbf{x}$ and $\mathbf{x}'$ represent the image points, in the left and right cameras, respectively.}
\end{itemize}

Furthermore, we assume that the world coordinate system coincides with 
the left camera coordinate system. The extrinsic parameters of the 
stereo system are the relative rotation between the two cameras 
$\mathbf{R}$ and, also, the relative translation $\mathbf{T}$. 
Therefore, $\mathbf{R}$ is the rotation matrix and $\mathbf{T}$ is the 
translation vector of the camera 2 relatively to camera 1. 

\subsection{Radial Distortion}
\label{radist}
The images acquired with this system are affected by radial 
distortion.  Radial distortion causes an inward or outward 
displacement of the pixels(relative to the center of distortion, which may or may 
not coincide with the image center--in this paper we assumed that 
they coincide), with the displacement being proportional to the radial 
distance. In general, radial distortion can be modeled by a 
polynomial model~\cite{article:Brown:1966}, by a division 
model~\cite{article:fitzgibbon:2001} or by modeling the camera as a 
non-central camera~\cite{miraldo,szeliski:2010}. For general parametric non-central models 
that use (for example) splines, 3D lines corresponding to neighboring pixels are 
taken into account in the model.
In general the polynomial model is 
applied to the metric coordinates of the pixels (i.e. before the 
transformation from the image to the sensor plane), while the 
division model can be applied either to the pixel coordinates in the 
image plane or in the sensor plane~\cite{scaramuzza:2006}. The 
polynomial model is used in Bouguet's Camera Calibration 
Toolbox~\cite{bouguet} and in Matlab's Computer Vision Toolbox. For 
the purpose of this paper, we considered both the polynomial and the 
division model.  

\subsubsection{Polynomial model}
Considering the rigid transformation between the world coordinate 
system and the 
camera coordinate system $\mathbf{T}_{W}^{C}$, then,  we have: 
\begin{equation}
  \mathbf{X}^C=^{C}\mathbf{T}_{W}\mathbf{X}^W
\end{equation}
where $\mathbf{X}^C={\left({X}^C,{Y}^C,{Z}^C\right)^{T}}$ stands 
for the point coordinates in the camera coordinate system, and 
$\mathbf{X}^W=\left({X}^W,{Y}^W,{Z}^W,1\right)^{T}$
stands for the homogeneous point coordinates in the world coordinate 
system. The normalized image coordinates are given by:
\begin{equation}
  \mathbf{x}_n=\left[\begin{array}{c}
      {X}^C/{Z}^C\\
      {Y}^C/{Z}^C
  \end{array} \right]
\end{equation}
where $\mathbf{x}_n={\left({x}_n,{y}_n\right)^{T}}$. The coordinates 
of the image points affected by radial distortion are obtained from 
(if we use a model with a polynomial of the $6^{th}$ degree):
\begin{equation}
  \mathbf{x}_d={\left(1 + {k}_1{r}^2+ {k}_2{r}^4+ 
  {k}_3{r}^6\right)}\left[\begin{array}{c}
      {x}_n\\
      {y}_n
  \end{array} \right]
\end{equation}
where ${k}_1$, ${k}_2$ and ${k}_3$ are the radial distortion 
parameters,  ${r}^2={x}_n^2+{y}_n^2$ is the radial distance,  and 
$\mathbf{x}_d={\left({x}_d,{y}_d\right)^{T}}$ are coordinates 
of the distorted points. After taking into account radial 
distortion, the pixel coordinates are computed from:
\begin{equation}
  \centering
  \label{eq:2Dprojection}
  \mathbf{x}=\mathbf{K}\left[\begin{array}{ccc}
      {x}_d &
      {y}_d &
      1
  \end{array} \right]^T
\end{equation}
where $\mathbf{K}$ is the matrix of the intrinsic parameters and 
$\mathbf{x}={\left({x},{y},1\right)^{T}}$ are the image 
coordinates, in pixels.

\subsubsection{Rational and Division models}
\label{division}

One of the difficulties with the application of the polynomial model 
is its inversion, since, in general, it is not analytically 
invertible. One possibility is to use the terms of the Taylor series 
expansion~\cite{whelan} or perform the numerical inversion. As an 
alternative it was proposed to model the distortion using either 
the rational function 
model~\cite{article:claus:2005} or the division 
model~\cite{article:fitzgibbon:2001}. As already mentioned, and for 
the purpose of modeling the stereo system described in this paper,
we used the division model. The 
parameters of the division model can be computed using either the 
fundamental matrix 
(~\cite{article:kukelova:2010,article:mivcuvsik:2003,article:fitzgibbon:2001})
or homographies 
(~\cite{article:fitzgibbon:2001,article:kukelova:2015}). This model 
was originally proposed 
for uncalibrated images, with a $2^{nd}$ order polynomial in the 
denominator, but it
can also be used with calibrated pixels, i.e. with normalized 
coordinates and with higher order polynomials~\cite{scaramuzza:2006}. 

This model allows the computation of the undistorted pixels
directly from the coordinates of the pixel with distortion, thereby avoiding the inversion difficulty of the polynomial 
model. Specifically we have:
\begin{equation}
  \centering
  \label{eq:divismodel}
  \mathbf{x}_u=\left[\begin{array}{c}
      {x}_u\\
      {y}_u\\
      1
  \end{array} \right]  =\left[\begin{array}{c}
      \frac{{x}_d}{1 + \lambda {r}_d^2}\\
      \frac{{y}_d}{1 + \lambda {r}_d^2}\\
      1
  \end{array} \right]=\left[\begin{array}{c}
      {x}_d\\
      {y}_d\\
      1 + \lambda {r}_d^2
  \end{array} \right]
\end{equation}
where $(x_u,y_u)$ are the coordinates of the undistorted pixel, 
$(x_d,y_d)$ are the coordinates of the distorted pixel,
${r}_d^2={x}_d^2+{y}_d^2$ is the radial distance, and $\lambda$ is the 
distortion coefficient. The division model can use higher order 
polynomials in the denominator, when additional distortion 
coefficients have to be estimated.

The computation of $\lambda$ can be performed using 
fundamental/essential matrices or homographies. In our case, and 
given that we used planar checkerboards for calibration, we used 
homographies to estimate the values of $\lambda$ for both cameras.
If we consider a plane whose images are acquired with  
projection matrices 
$\mathbf{P}=\left[\mathrm{I}\,|\,0\right]$ and 
$\mathbf{P}'=\left[\mathrm{R}\,|\,{t}\right]$, the relationship 
between the images of the points that belong to the plane is given by (up 
to a scale factor)~\cite{article:kukelova:2015}:
\begin{equation}
  \centering
  \label{eq:homography}
  \mathbf{x}^{'}_{i} = \mathbf{H}\mathbf{x}_i
\end{equation}
where $\mathbf{x}'_i={\left({x}'_i,{y}'_i,1\right)^{T}}$ and 
$\mathbf{x}_i={\left({x}_i,{y}_i,1\right)^{T}}$ represent the 
coordinates of the $i^{th}$ corresponding pixels and 
$\mathbf{H}\in\mathbb{R}^{3\times 3}$ is the homography matrix 
(\cite{hartley:2000}). This relationship between the images of 
the plane is used to compute the radial distortion parameters for 
each of the two cameras (corresponding to the division model with 
only one parameter, ie.~a second order polynomial). Let us denote 
$\lambda$ as the parameter for the first camera and $\lambda'$ for 
the second camera.

Then, multiplying, on the left, Eq.~\ref{eq:homography} by the skew 
symmetric matrix corresponding to $\mathbf{x}^{'}$ we obtain:
\begin{equation}
  \centering
  \label{eq:solve:homography}
  \left[\begin{array}{ccc}
      0 & -{w}'_i & {y}'_i\\
      {w}'_i & 0 & -{x}'_i\\
      -{y}'_i & {x}'_i & 0
  \end{array} \right] \left[\begin{array}{ccc}
      {h}_{11} & {h}_{12} & {h}_{13}\\
      {h}_{21} & {h}_{22} & {h}_{23}\\
      {h}_{31} & {h}_{32} & {h}_{33}
  \end{array} \right]\left[\begin{array}{c}
      {x}_i\\
      {y}_i\\
      {w}_i
  \end{array} \right]=\mathbf{0}
\end{equation}
where ${w}_i=1 + \lambda {r}_d^2$ and ${w}'_i=1 + \lambda' 
{r}_d^{'2}$.

From the computation of the planar homography~\cite{article:kukelova:2015}) both distortion coefficients can be 
estimated. The knowledge of $\lambda$ and $\lambda'$ allows the 
estimation of the undistorted pixels, using the 
division model (Eq.~\ref{eq:divismodel}).

\begin{figure}[t]
  \centering
  \includegraphics[width=0.23\textwidth]{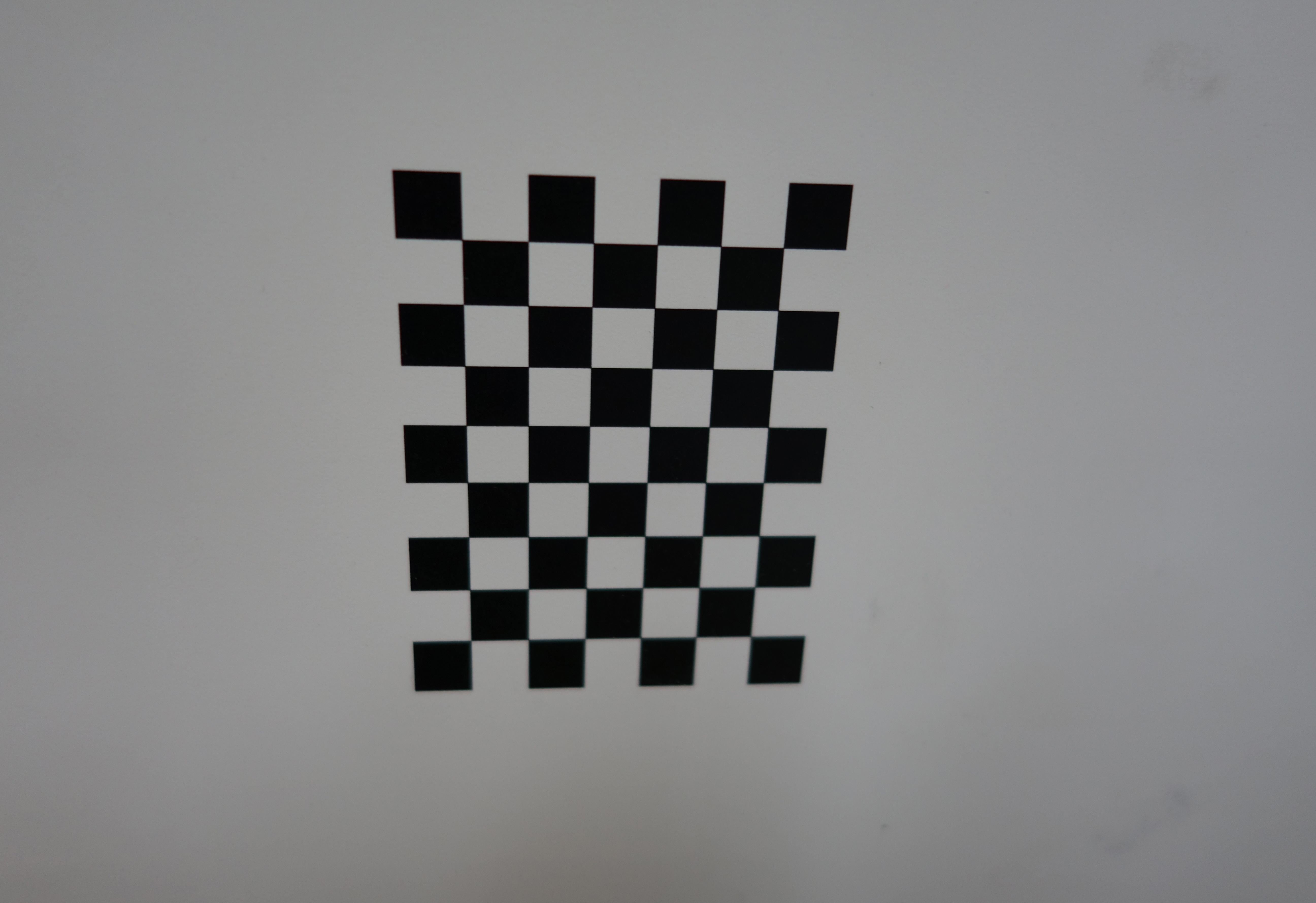}\hfill
  \includegraphics[width=0.23\textwidth]{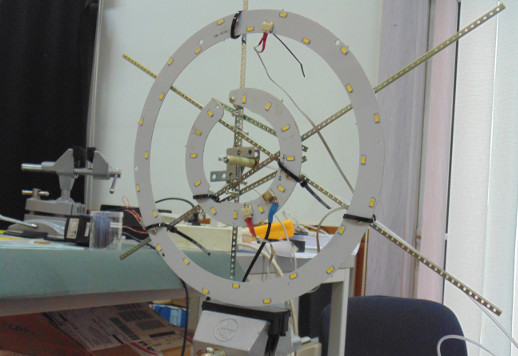}
  \caption{Image of the experimental setup and of one of the checkerboards used in the calibration.}
  \label{Fig:checker_1}
\end{figure}

\subsubsection{Non-Central Model}
Non-central models can also be used to model distortion (radial and 
tangential). One possibility is to use the model proposed and 
described in~\cite{miraldo}. This kind of model can represent generic 
types of distortions. 

\section{Stereo System Calibration}\label{calib}
To perform the calibration of the stereo system, we used images of a 
checkerboard, made up of black squares (see Fig.~\ref{Fig:checker_1}). Bouguet's calibration 
toolbox~\cite{bouguet} was used for that purpose. In the specific 
case of the cameras that we are dealing with, and as a result of the 
image low resolution and  quality, the uncertainties on the estimated 
parameters depend on the sizes of the checkerboard squares and, 
therefore, on the average of the distances of the checkerboards from 
the cameras. For the purpose of calibration, 
three sets of nineteen images were used. For each set, a different checkerboard was used (squares with different dimensions). All the checkerboards were 
made up of 9 rows and 7 columns of black and white
squares. From each image 48 points were used. An image 
of the acquisition setup is presented 
in Fig.~\ref{Fig:checker_1}. The dimensions of the squares and 
average distances from the cameras were:
\begin{description}
  \item [Dataset 1:] Squares with side lengths of 2.25 mm,
    located at an average distance of 20 mm from the cameras;
  \item [Dataset 2:] Squares with side lengths of 4.50 mm,
    located at an average distance of 50 mm from the cameras; and
  \item [Dataset 3:] Squares with side lengths of 5.99 mm,
    located at an average distance of 70 mm from the cameras.
\end{description}

Three calibrations were performed and the intrinsic parameters for 
both cameras are presented in Tabs.~\ref{tabcalib:1} 
and~\ref{tabcalib:2}. The intrinsic parameters are:
\begin{itemize}
  \item {$f_x$ and $f_y$:} Focal lengths in pixels along the 
    $x$ and $y$ axes;
  \item {$u_0$ and $v_0$:} Coordinates of the principal point; and
  \item {$k_1, k_2, k_3$:} Radial distortion parameters 
    corresponding, respectively to $r^2, r^4, r^6$.
\end{itemize}

%
\begin{table}[t]
\centering
\small{
  \begin{tabular}{l|c|c|c|}
    \cline{2-4}
    & \multicolumn{3}{c|}{Camera 1/Left Camera} \\ \hline
    \multicolumn{1}{|l|}{}	& Dataset 1 & Dataset 2 & Dataset 3 \\ \hline
    \multicolumn{1}{|l|}{${f}_x ({px})$}	& 216.520$\pm$1.399  & 
    216.307$\pm$1.898 & 216.360$\pm$0.579
    \\  \hline
    \multicolumn{1}{|l|}{${f}_y ({px})$}	& 216.259$\pm$1.448 & 
    216.421$\pm$1.947 & 216.315$\pm$0.576 
    \\  \hline
    \multicolumn{1}{|l|}{${u}_0 ({px})$ }	& 122.241$\pm$1.106 & 
    121.847$\pm$1.162 & 122.414$\pm$0.758
    \\   \hline
    \multicolumn{1}{|l|}{${v}_0 ({px})$ }	& 113.410$\pm$0.826  & 
    112.981$\pm$0.970  & 111.535$\pm$0.646 \\ \hline
    \multicolumn{1}{|l|}{$k_1$ }	& -0.349$\pm$0.010 & -0.353$\pm$0.011  & -0.369$\pm$0.023  \\ \hline
    \multicolumn{1}{|l|}{$k_2$ }	& 0.148$\pm$0.021 & 0.144$\pm$0.021  & 0.303$\pm$0.182  \\ \hline
    \multicolumn{1}{|l|}{$k_3$ }	& 0.000$\pm$0.000 & 0.000$\pm$0.000  & -0.366$\pm$0.417  \\ \hline
  \end{tabular}} 
  \caption{Intrinsic parameters for the left camera.}
  \label{tabcalib:1}
\end{table}

\begin{table}[t]
\centering
\small{
  \begin{tabular}{l|c|c|c|}
    \cline{2-4}
    & \multicolumn{3}{c|}{Camera 2/Right Camera} \\ \hline
    \multicolumn{1}{|l|}{}	& Dataset 1 & Dataset 2 & Dataset 3 
    \\ \hline
    \multicolumn{1}{|l|}{${f}'_x ({px})$}	& 213.860$\pm$1.903  
    & 
    215.601$\pm$2.112 & 216.258$\pm$0.585
    \\  \hline
    \multicolumn{1}{|l|}{${f}'_y ({px})$}	& 213.325$\pm$1.943 & 
    215.412$\pm$2.161 & 216.379$\pm$0.583 
    \\  \hline
    \multicolumn{1}{|l|}{${u}'_0 ({px})$ }	& 123.132$\pm$1.246 & 
    123.486$\pm$1.278 & 124.419$\pm$0.739
    \\   \hline
    \multicolumn{1}{|l|}{${v}'_0 ({px})$ }	& 124.568$\pm$0.906  
    & 
    122.412$\pm$1.084  & 122.283$\pm$0.635 \\ \hline
    \multicolumn{1}{|l|}{$k'_1$ }	& -0.344$\pm$0.011 & -0.342$\pm$0.012  & -0.336$\pm$0.004  \\ \hline
    \multicolumn{1}{|l|}{$k'_2$ }	& 0.143$\pm$0.017 & 0.124$\pm$0.020  & -0.103$\pm$0.208  \\ \hline
    \multicolumn{1}{|l|}{$k'_3$ }	& 0.000$\pm$0.000 & 0.000$\pm$0.000  &  0.866$\pm$0.499  \\ \hline
  \end{tabular} }
  \caption{Intrinsic parameters for the right camera.}
  \label{tabcalib:2}
\end{table}

Given the three calibrations, and taking into account the 
estimated uncertainties on the parameters, we decided to use the 
values corresponding to the calibration performed with the 
\textbf{Dataset 3}. From these experiments, it is clear that the images 
acquired by both cameras are strongly affected by radial distortion.
This results from the fact that the coefficients of $r^6$, where $r$ is the radial distance, are 
non-negligible. In addition to the intrinsic parameters for both 
cameras of the stereo system, we also estimated the relative pose 
between both cameras. This stereo pair is designed and built so that 
its configuration can be represented by a fronto-parallel system. However 
inaccuracies occur during the 
manufacture process. Moreover, since the camera dimensions are extremely small, 
small deviations from the nominal values may affect the 
reconstruction accuracy.


\begin{table}[t]
\centering
\small{
  \begin{tabular}{|l|c|}
    \hline
    Relative Pose	&   \\ \hline \hline
    $ \omega_x (rad)$ &  $-0.0034$  \\ \hline
    $ \omega_y (rad)$  &  $0.0134$  \\ \hline
    $ \omega_z (rad)$  &  $0.0341$   \\ \hline
    $ 	T_x (mm)$  &  $1.063 \pm 0.040$  \\ \hline
    $T_y (mm)$  &    $-0.203\pm0.038$   \\ \hline
    $T_z (mm)$   &  $0.118 \pm 0.153$  \\ \hline
  \end{tabular}}
  \caption{Relative pose.}
  \label{tab:pose}
\end{table}

The values for relative pose between the two cameras are shown in 
Tab.~\ref{tab:pose}. The reference coordinate system is the 
coordinate system of the left camera. Rotation is represented by the 
angle-rotation values (Rodrigues' representation of rotation) and specifies the  
rotation between the right and left 
camera coordinate systems. From the results, we concluded that the rotation can be represented by
an angle of approximately $2.1^{\degree}$, around a rotation axis. The 
translation vector specifies the coordinates of the origin of right 
camera coordinate system in the left camera coordinate system. As 
it can be seen from those values, the stereo pair slightly deviates 
from the fronto-parallel configuration. The world coordinate system 
coincides with the left coordinate system, and the 3D reconstruction 
will be performed in the reference coordinate system of the left 
camera.

\begin{table}[t]
\centering
\small{
  \begin{tabular}{|l|c|c|c|} \hline 
    & P-average  (mm)  & D-average (mm) & R (pixels) \\ \hline \hline
    Plane 1 & $2.718$   & $1.267$ &  $0.077$    \\ \hline
    Plane 2 & $3.497$   &   $3.2350$     &  $0.119$   \\ \hline  
    Plane 3 & $2.805$   &   $3.280$  &    $0.110$       \\  \hline
    Plane 4 & $1.777$   &   $0.1460$  &  $0.313$  \\ \hline
    Plane 5	& $5.702$   &  $5.7960$  &  $0.180$  \\ \hline
  \end{tabular}}
  \caption{Results for the baseline approach.}
  \label{tab:refres}
\end{table}

\section{Experimental Results}

In this section, firstly, we define the criteria used for the evaluation, Sec.\ref{sec:evaluation_criteria}.
Then we tested different reconstruction methods, Sec.~\ref{sec:reconstruction_methods}.

\subsection{Evaluation Criteria}\label{sec:evaluation_criteria}
To evaluate the 3D reconstruction results three error criteria were 
used:
\begin{description}
  \item [P-Planarity:] The 3D points to be reconstructed 
    lie on planes. Therefore, one of the criteria used to evaluate 
    the quality of the reconstruction is its planarity.
    For that purpose, for each set of reconstructed 
    points belonging to a plane, a 3D plane was estimated using 
    least-squares. Next, the distance between the 3D 
    reconstructed points and the reconstructed plane was estimated.
    We repeat this procedure for all the points, and get the
    average value of the estimated distances. A perfectly 
    planar reconstruction would correspond to a zero average 
    distance (per plane);
  \item [D-3D distance:] Since the 3D reconstructed points 
    correspond to corners of the squares located in planes, the ground truth 3D 
    distances between consecutive 3D points are known. The distances 
    between consecutive 3D points were measured and the 
    deviations relative to the ground truth (errors) are computed. 
    Their average values (per plane) characterize also the quality of 
    3D reconstruction; and
  \item [R-Reprojection error:] The re-projection error is 
    obtained by re-projecting the 3D reconstructed points onto the 
    images and measuring the distances relative to their 
    ground truth images.
\end{description}

\subsection{Reconstruction Methods}\label{sec:reconstruction_methods}
To evaluate the 3D reconstruction algorithms, we defined a 
baseline method using the algorithms available in the 
Bouguet and Matlab Computer Vision toolboxes. For that
purpose, the following methods were applied to obtain a reference 3D 
reconstruction:
\begin{description}
  \item [1-Undistort:] The image coordinates of the points 
    were undistorted by inverting the polynomial distortion model, using 
    a numeric non-linear least-squares optimization; and
  \item [2-Triangulate:] Using the coordinates of 
    corresponding undistorted image points, the intrinsic parameters 
    of both 
    cameras and the relative pose the coordinates of the 3D points 
    were estimated.
\end{description}

All nineteen images were used to evaluate the reconstruction. 
However, and for reasons of space, five were chosen to display the 
results of the evaluation. In each one of the five images, 48 points 
were 
used. 
The reference distance between the 3D points is 
$5.99$ mm.



The results of the evaluation, for the three error criteria 
previously described, are presented in Tab.~\ref{tab:refres}.

As mentioned in Sec.~\ref{radist}, the polynomial model requires a 
numerical inversion. For that reason, the division model was 
proposed~\cite{article:fitzgibbon:2001, article:kukelova:2015, 
article:kukelova:2010} since it allows the computation of the 
undistorted coordinates directly from the distorted coordinates, either from 
the uncalibrated or calibrated images. Using the homography based 
approach described in Sec.~\ref{division}, we computed the 
distortion parameters for the division model. We computed a single 
parameter for each image (i.e. we used a second-order polynomial in 
the denominator). For the computation of the distortion parameters 
and for the un-distortion of the images, we considered that the 
center 
of distortion coincided with the center of the image. Reconstruction 
was performed using triangulation 
and plane-sweeping~\cite{sweep07,sweep96}, after the un-distortion of 
the images and using coordinates in pixels. The plane-sweeping method 
was applied naively, in a direct way. However we plan to apply it 
adaptively, as a function of the distortion and region of space to be 
reconstructed. The results, for the same 
planes, are presented in Tabs.~\ref{tab:divtri} 
and~\ref{tab:divsweep}. 

\begin{table}[t]
\centering
\small{
  \begin{tabular}{|l|c|c|c|} \hline 
    & P-average  (mm)  & D-average (mm) & R (pixels) \\ \hline 
    \hline
    Plane 1 & $5.806$   & $2.013 $ &  $1.512$    \\ \hline
    Plane 2 & $5.158$   &   $ 4.128$     &  $1.215$   \\ \hline  
    Plane 3 & $3.005$   &   $ 3.494$  &    $0.405$     \\  
    \hline
    Plane 4 & $2.482$   &   $0.489 $  &  $1.171$  \\ \hline
    Plane 5	& $7.629$   &  $14.041 $  &  $2.117$  \\ \hline
  \end{tabular}}
  \caption{Results for the division model with triangulation.}
  \label{tab:divtri}
\end{table}

\begin{table}[t]
\centering
\small{
  \begin{tabular}{|l|c|c|c|} \hline 
    & P-average  (mm)  & D-average (mm) & R (pixels) \\ \hline 
    \hline
    Plane 1 & $5.802$   & $2.015 $ &  $1.512$    \\ \hline
    Plane 2 & $5.166$   &   $4.117 $     &  $1.215$   \\ \hline  
    Plane 3 & $3.002$   &   $3.481 $  &    $0.405$       \\  
    \hline
    Plane 4 & $2.479$   &   $0.483 $  &  $1.171$  \\ \hline
    Plane 5	& $7.632$   &  $14.022 $  &  $2.117$  \\ \hline
  \end{tabular}}
  \caption{Results for the division model with plane-sweep.}
  \label{tab:divsweep}
\end{table}

These results show that the plane-sweeping approach (as it was 
applied) is equivalent to triangulation. However, when comparing the 
removal of 
distortion using the division model with the baseline approach, it 
clearly performs worse. The main reason for this behavior stems from 
the fact that, for each camera, only a single distortion parameter is used.

\begin{table}[t]
\centering
\small{
  \begin{tabular}{|l|c|c|c|} \hline 
    & P-average  (mm)  & D-average (mm) & R (pixels) \\ 
    \hline 
    \hline
    Plane 1 & $6.942$   & $0.8080 $ &  $2.716$    \\ \hline
    Plane 2 & $4.206$   &   $3.7432 $     &  $0.854$   \\ 
    \hline  
    Plane 3 & $4.864$   &   $4.5740 $  &    $1.656$       \\  
    \hline
    Plane 4 & $1.640$   &   $0.1930 $  &  $0.353$  \\ \hline
    Plane 5	& $833.587$   &  $1482 $  &  $17.741$  \\ \hline
  \end{tabular}}
  \caption{Results for the division model with normalized 
  coordinates, un-distortion and then triangulation.}
  \label{tab:divnormtri}
\end{table}

\begin{table}[t]
\centering
\small{
  \begin{tabular}{|l|c|c|c|} \hline 
    & P-average  (mm)  & D-average (mm) & R (pixels) \\ 
    \hline 
    \hline
    Plane 1 & $ 5.4793$   & $2.015 $ &  $2.720$    \\ \hline
    Plane 2 & $4.070$   &   $4.1170 $     &  $0.908$   \\ 
    \hline  
    Plane 3 & $4.433 $   &   $3.4810 $  &    $1.710$       
    \\  
    \hline
    Plane 4 & $1.594 $   &   $0.9110 $  &  $0.416$  \\ \hline
    Plane 5	& $26.554$   &  $42.788 $  &  $18.102$  \\ \hline
  \end{tabular}}
  \caption{Results for the division model with normalized 
  coordinates and the triangulating with distorted coordinates 
  Eq.~\ref{eq_distort}.}
  \label{tab:divnormtri2}
\end{table}

\subsubsection{Center of Distortion}
To evaluate the importance of the center of distortion, we applied the 
distortion model to the normalized images, i.e. After multiplying 
the image by the inverse of the intrinsic parameters' matrix. 
In the case of the NanEye stereo cameras, the principal 
points differ from the center of the images (see 
Sec.~\ref{calib}). New values for the radial distortions 
parameters (division model) were computed, also using 
homographies. A triangulation method that directly estimated the 
depths, using the distortion parameters,
was applied. In this case, reconstruction was performed in a single 
step. The method performs triangulation directly with the distorted 
points (with known distortion parameters).  Let 
$\mathbf{x_{d}^{'}}$ be the normalized distorted coordinates in the 
right image and $\mathbf{x_{d}}$ normalized distorted coordinates in 
the left image with:
\begin{equation}
  \centering
  \mathbf{x}'_d=\left[\begin{array}{ccc}
      {x}'_d &
      {y}'_d &
      1 + \lambda' {r}_d^2
  \end{array} \right]^T
\end{equation}	
and 
\begin{equation}
  \centering
  \mathbf{x}_d=\left[\begin{array}{ccc}
      {x}_d &
      {y}_d &
      1 + \lambda {r}_d^2
  \end{array} \right]^T
\end{equation}		

The relationship between the right and left distorted
normalized coordinates is given by
\begin{equation}
  \alpha_1\mathbf{x}'_d=\alpha_2\mathbf{R}\mathbf{x}_d + \mathbf{T}
  \label{eq_distort}
\end{equation}
where $\alpha_1$ and $\alpha_2$ are the scale factors and 
$\mathbf{R}$ and $\mathbf{T}$ are the rotation matrix 
and translation vector, respectively.

Let $\widehat{\mathbf{x}}'_d $ be the skew-symmetric matrix 
corresponding to vector $\mathbf{x_{d}^{'}}$. Then, multiplying on 
the left the previous equation by $\widehat{\mathbf{x}}'_d $, we 
obtain:
\begin{equation}
  \alpha_2 \widehat{\mathbf{x}}'_d \mathbf{R}\mathbf{x}_d + 
  \widehat{\mathbf{x}}'_d\mathbf{T}=\mathbf{0}
\end{equation}
Using this constraint, a system of equations, for all corresponding 
points in the left and right images, can be determined. The solution 
of this system of equations can be computed by means of SVD, yielding the $\alpha_2$ for 
all points, which constitutes their depths.

The results obtained by first un-distorting the points and then 
triangulating are presented 
in Tab.~\ref{tab:divnormtri}, while the results with
triangulating directly with the distorted points (using estimates of 
the distortion, Eq.~\ref{eq_distort}) are represented in 
Tab.~\ref{tab:divnormtri2}. 

These results show that the application of the distortion model 
requires a good estimate of the center of distortion. In the case of 
these cameras, the principal point differs from the center of 
distortion, which is one of the reasons for these results. 
Furthermore the results with the triangulation with the distorted 
coordinates are equivalent to first un-distorting the points and then 
triangulating.

\subsubsection{Distortion Models: Comparison}
To further show that the division model (with one parameter) is 
insufficient to represent the radial distortion that is parametrized by the
polynomial model with three parameters, the following comparison was performed:
\begin{itemize}
  \item We defined six synthetic 3D points, with known 3D 
    coordinates and a distance of $2.189$ mm between neighboring 
    points;
  \item Using the estimated camera intrinsic parameters, the 
    polynomial distortion coefficients and the relative pose we 
    projected the points into the images; and
  \item Using these images, three reconstruction methods were performed: the baseline method; the
    division model \& triangulation; and the division model \& plane sweeping.
\end{itemize}

The results are presented in Tab.~\ref{tab:comp}. These results show 
that the division model with a single coefficient can not correctly 
model this level of distortion.

\begin{table}[t]
\centering
\small{
  \begin{tabular}{|l|c|c|c|} \hline
    & P-average  (mm)  & D-average (mm) & R (pixels) \\ 
     \hline \hline
    Baseline & $0.256$ & $0.6280$ & $0.863$ \\ \hline
    Division Triang & $0.008$ & $9.291$ & $57.556$ \\ \hline
    Division Sweep  & $0.006$ & $9.283$ & $59.374$ \\ \hline
  \end{tabular}}
  \caption{Comparison of distortion models.}
  \label{tab:comp}
\end{table}

\begin{table}[t]
  \centering
  {\small
  \begin{tabular}{l|c|c|c|}
    \cline{2-4}
    & \multicolumn{3}{c|}{Camera 1/Left Camera} \\ \hline
    \multicolumn{1}{|l|}{}	& Test 1 & Test 2 & Test 3 \\ \hline
    \multicolumn{1}{|l|}{${f}_x ({px})$}	& 217.906$\pm$1.009  & 216.332$\pm$0.802 & 217.396$\pm$0.996\\  \hline
    \multicolumn{1}{|l|}{${f}_y ({px})$}	& 217.855$\pm$1.004 &  216.187$\pm$0.827 & 217.121$\pm$0.950 \\  \hline
    \multicolumn{1}{|l|}{${u}_0 ({px})$ }	& 121.501$\pm$1.096 & 123.041$\pm$1.044 & 121.813$\pm$1.055\\   \hline
    \multicolumn{1}{|l|}{${v}_0 ({px})$ }	& 113.061$\pm$0.954  & 111.483$\pm$0.851  & 113.324$\pm$0.954 \\ \hline
    \multicolumn{1}{|l|}{$k_1$ }	& -0.379$\pm$0.034 & -0.383$\pm$0.029  & -0.376$\pm$0.034  \\ \hline
    \multicolumn{1}{|l|}{$k_2$ }	& 0.384$\pm$0.276 & 0.370$\pm$0.220  & 0.361$\pm$0.182  \\ \hline
    \multicolumn{1}{|l|}{$k_3$ }	& -0.588$\pm$0.663 & -0.425$\pm$0.480  & -0.575$\pm$0.658  \\ \hline
  \end{tabular} }
  \caption{Intrinsic parameters for the left camera.}
  
  \label{tabcalib:2_1}
\end{table}

\subsection{Evaluation Based on Cross-Validation}

As it was mentioned in Sec.~\ref{calib}, we used the calibration
obtained using the Dataset 3 because of the low uncertainties 
obtained in the estimated parameters. For that 
calibration we used all 19 images to calibrate and, then, used  
images from the same set to evaluate the reconstruction.
Using a subset of the calibration images to evaluate the 
reconstruction may lead to over-fitting. Therefore we decided to use 
cross-validation to evaluate the reconstruction.
Thus, we performed  new tests using sets of 10 images for 
calibration, and the remaining 9 images (in each set) to evaluate the 
3D reconstruction. Three tests 
were made using different combinations/sets  of images. Additionally, 
we used a RANSAC-based
plane fitting method to reduce the number of outliers and to improve 
the
planarity results.  We used as threshold 15 per cent of the distance 
between the plane and the origin.
Also, a median filter was used to decrease the error due to
outliers in 3D distance error.


As it was described in Sec.~\ref{calib}, we used the calibration 
obtained using the Dataset 3 because of the low uncertainties
given by the calibration procedure. For this calibration
we used 19 images to calibrate and  a subset of the
 images to evaluate the reconstruction.


The results for the calibration are presented in Tabs.~\ref{tabcalib:2_1}~and~\ref{tabcalib:2_2}.

\begin{table}[t]
  \centering	
  {\small
  \begin{tabular}{l|c|c|c|}
    \cline{2-4}
    & \multicolumn{3}{c|}{Camera 2/Right Camera} \\ \hline
    \multicolumn{1}{|l|}{}	& Test 1 & Test 2 & Test 3 \\ \hline
    \multicolumn{1}{|l|}{${f}'_x ({px})$}	& 215.866$\pm$0.993  & 216.837$\pm$0.812 & 215.473$\pm$0.987\\  \hline
    \multicolumn{1}{|l|}{${f}'_y ({px})$}	& 215.902$\pm$0.992 &  217.131$\pm$0.839 & 215.792$\pm$0.947 \\  \hline
    \multicolumn{1}{|l|}{${u}'_0 ({px})$ }	& 126.170$\pm$1.062 & 125.429$\pm$1.042 & 125.270$\pm$1.012\\   \hline
    \multicolumn{1}{|l|}{${v}'_0 ({px})$ }	& 120.638$\pm$0.925  & 122.122$\pm$0.847  & 121.327$\pm$0.917 \\ \hline
    \multicolumn{1}{|l|}{$k'_1$ }	& -0.330$\pm$0.032 & -0.333$\pm$0.032  & -0.316$\pm$0.034  \\ \hline
    \multicolumn{1}{|l|}{$k'_2$ }	& -0.223$\pm$0.257 & -0.078$\pm$0.262  & -0.281$\pm$0.271  \\ \hline
    \multicolumn{1}{|l|}{$k'_3$ }	& 1.228$\pm$0.605 & 0.612$\pm$0.612  & 1.356$\pm$0.640  \\ \hline
  \end{tabular} 
  }
  \caption{Intrinsic parameters for the right camera.}
  \label{tabcalib:2_2}
\end{table}

\begin{table}[t]
\centering
  {\small
    \begin{tabular}{|c|l|c|c|c|c|} 
      \hline 
      & & P-average  (mm)  & D-average (mm) & R (pixels) \\ 
      \hline \hline 
      \parbox[t]{2mm}{\multirow{3}{*}{\rotatebox[origin=c]{90}{Test 1}}} & Plane A1 & $2.070$   & $1.172 $ 	&  $0.114$  	  \\ \cline{2-5}     
      & Plane B1 & $3.046$   &   $3.000$  	&  $0.114$  	\\ \cline{2-5}
      & Plane C1 & $2.323$   &  $1.209$  	&  $0.417$ 	\\ \hline

      \parbox[t]{2mm}{\multirow{3}{*}{\rotatebox[origin=c]{90}{Test 2}}} & Plane A2 & $2.182$   &  $2.401$    &  $0.172$  	 \\ \cline{2-5}  
      & Plane B2 & $2.857$   &  $5.011$  	&  $0.245$ 	 \\ \cline{2-5}
      & Plane C2 & $2.760$   &  $1.380$  	&  $0.455$  	\\ \hline

      \parbox[t]{2mm}{\multirow{3}{*}{\rotatebox[origin=c]{90}{Test 3}}} & Plane A3 & $2.818$   &  $2.216$    &  $0.059$  	 \\ \cline{2-5}  
      & Plane B3 & $3.611$   &  $3.888$  	&  $0.088$ 	 \\ \cline{2-5}
      & Plane C3 & $3.112$   &  $3.536$  	&  $0.257$  	\\ \hline
    \end{tabular}
  }
  \caption{Results for the baseline model.}
  \label{tab:resul_final1}
\end{table}

\begin{table}[t]
\centering
  {\small
    \begin{tabular}{|c|l|c|c|c|c|} 
      \hline 
    & & P-average  (mm)  & D-average (mm) & R (pixels) \\ 
    \hline \hline
    \parbox[t]{2mm}{\multirow{3}{*}{\rotatebox[origin=c]{90}{Test 1}}} & Plane A1 & $2.740$   & $1.489$ 	&  $1.792$  	 \\ \cline{2-5} 
    & Plane B1 & $2.682$   &   $3.063$  	&  $0.872$  \\ \cline{2-5} 
    & Plane C1 & $2.004$   &  $1.198$  	&  $1.240$  \\ \hline

    \parbox[t]{2mm}{\multirow{3}{*}{\rotatebox[origin=c]{90}{Test 2}}} & Plane A2 & $3.052$   & $3.503$ 	&  $1.895$    \\ \cline{2-5}
    & Plane B2 & $4.547$   & $7.225$  	&  $2.356$  \\ \cline{2-5}
    & Plane C2 & $2.747$   & $2.738$  	&  $1.227$  \\ \hline

    \parbox[t]{2mm}{\multirow{3}{*}{\rotatebox[origin=c]{90}{Test 2}}} & Plane A3 & $3.273$   & $2.842 $ &  $1.495$    \\ \cline{2-5}
    & Plane B3 & $2.986$   & $4.301 $  &  $0.743$  \\ \cline{2-5}
    & Plane C3 & $2.793$   & $2.728 $  &  $1.593$ \\ \hline
  \end{tabular}}
  \caption{Results for the division model with triangulation.}
  \label{tab:resul_final2}
\end{table}

\begin{table}[t]
\centering
  {\small
      \begin{tabular}{|c|l|c|c|c|c|} 
      \hline 
    & & P-average  (mm)  & D-average (mm) & R (pixels) \\ 
    \hline \hline
    \parbox[t]{2mm}{\multirow{3}{*}{\rotatebox[origin=c]{90}{Test 1}}} & Plane A1 & $2.578$   & $1.862 $ &  $4.497$   \\ \cline{2-5} 
    & Plane B1 & $2.365$   &   $2.946 $  &  $0.431$  \\ \cline{2-5}
    & Plane C1 & $2.077$   &  $1.283 $  &  $0.531$  \\ \hline

    \parbox[t]{2mm}{\multirow{3}{*}{\rotatebox[origin=c]{90}{Test 2}}} & Plane A2 & $4.521$   & $2.899$ 	&  $2.474$    \\ \cline{2-5} 
    & Plane B2 & $3.928$   & $38.602$  	&  $9.103$  \\ \cline{2-5}
    & Plane C2 & $2.563$   & $2.757$  	&  $0.602$  \\ \hline

    \parbox[t]{2mm}{\multirow{3}{*}{\rotatebox[origin=c]{90}{Test 3}}} & Plane A3 & $2.396$   & $3.646 $ &  $2.961$   \\ \cline{2-5} 
    & Plane B3 & $3.690$   & $3.892 $  &  $0.705$  \\ \cline{2-5}
    & Plane C3 & $3.269$   & $4.054 $  &  $1.897$  \\ \hline

  \end{tabular}}
  \caption{Results for the division model with normalized coordinates and triangulation.}
  \label{tab:resul_final3}
\end{table}

For this experiment, the results presented in Tabs.~\ref{tab:resul_final1},\ref{tab:resul_final2} and \ref{tab:resul_final3} essentially confirm the conclusions previously drawn: 
the baseline method, with three radial distortion parameters 
generates better results than the variants of the division model we 
considered (with only one radial distortion parameter). The results 
for the baseline method are consistently better across all the three 
criteria, with only a few exceptions.

\section{Conclusions}
In this paper we fully characterize the NanEye stereo pair 
manufactured by CMOSIS/AWAIBA. We show that these cameras are 
affected by significant radial distortion. We used a polynomial of 
degree six to model this distortion.  We also show that the 
reconstruction  errors are significant and that reconstruction with 
higher accuracy 
requires both different radial distortion models and non-linear 
optimization methods. We also evaluated the division model with a 
single distortion coefficient and concluded that, while being simpler 
to use, it yields worse results. We plan to address the problem of 3D 
reconstruction with these cameras by jointly considering the radial 
distortion and the small baseline.



%
\bibliographystyle{abbrv}
\bibliography{sigproc-stereo}  
\end{document}